# Review of Machine Learning Methods for Additive Manufacturing of Functionally Graded Materials

*Mohammad Karimzadeh[1], Deekshith Basvoju[2], Aleksandar Vakanski[2], Indrajit Charit[2], Fei Xu[3], Xinchang Zhang[3]*

[1] Department of Computer Science, University of Idaho, Moscow, ID 83844, USA

[2] Department of Nuclear Engineering and Industrial Management, University of Idaho, Idaho Falls, ID 83402, USA

[3] Idaho National Laboratory, Idaho Falls, ID 83415, USA

**Abstract**

Additive Manufacturing (AM) is a transformative manufacturing technology enabling direct fabrication of complex parts layer-be-layer from 3D modeling data. Among AM applications, the fabrication of Functionally Graded Materials (FGMs) has significant importance due to the potential to enhance component performance across several industries. FGMs are manufactured with a gradient composition transition between dissimilar materials, enabling the design of new materials with location-dependent mechanical and physical properties. This study presents a comprehensive review of published literature pertaining to the implementation of Machine Learning (ML) techniques in AM, with an emphasis on ML-based methods for optimizing FGMs fabrication processes. Through an extensive survey of the literature, this review article explores the role of ML in addressing the inherent challenges in FGMs fabrication and encompasses parameter optimization, defect detection, and real-time monitoring. The article also provides a discussion of future research directions and challenges in employing ML-based methods in AM fabrication of FGMs.

1. **INTRODUCTION**

Additive Manufacturing (AM) refers to a collection of manufacturing processes where materials are directly joined to manufacture freeform parts layer-by-layer from a 3D computer-aided design model [1–6]. This technology has revolutionized the manufacturing of complex parts by enabling direct material printing and offers several advantages such as cost-effectiveness, manufacturing waste reduction, and opening new possibilities for manufacturing automation. AM also enables mass part customization and eliminates the need for hard-tooling of machining setups, making it suitable for on-demand manufacturing, thus mitigating supply chain challenges [2–4]. Nevertheless, AM also presents certain challenges. These encompass a lack of inherent repeatability and a shortage of widespread design knowledge and tools [3, 7, 8]. Also, another challenge in the acceptance of AM components is the lack of a streamlined process qualification methodology [8]. Additionally, the quality of AM-produced parts can be lower in comparison to conventional manufacturing methods, since defects negatively influence the structural integrity of the parts, due to the complex physics of AM processes [9, 10].

To fully leverage the benefits of AM, the design, processing, and production have become more complex in recent years. Accordingly, these complexities require significant knowledge for selecting and optimization of the AM process parameters. Consequently, although crucial for achieving high-quality products and minimizing material and time losses, the selection of process parameters can be time-consuming and expensive. Moreover, compositional inconsistencies in AM components arise due to the complex physics and the need for knowledge-supported practices to avoid defects [10–12].

One group of materials for which AM holds great potential for enhancing component properties is Functionally Graded Materials (FGMs). FGMs are advanced composite materials that are manufactured with a gradient composition transition between dissimilar materials, resulting in location-dependent mechanical and physical material properties [13, 14]. The composition gradient in FGMs enables enhancing the material properties by combining the advantages of different materials in a single component. Such material composition differs from traditional composite materials, characterized by a sharp inter-face between the matrix and reinforcement material. Additionally, the gradient transition between dissimilar materials in FGMs provides for location-wise properties and improved overall performance, making them desirable for aerospace, automobile, biomedical, and defense industries [15, 16]. Examples of FGMs include Stainless Steel 316/Inconel 718 [16] and Copper/Tungsten [17] FGMs for structural steels, Nickel/Aluminum Oxide FGM [18] for aerospace applications, Yttria-Stabilized Zirconia/Nickel-based superalloy FGM [19] for thermal barrier coatings, and Cobalt/Chromium [15] and Carbon Fiber Polymer Matrix/Yttria-Stabilized Zirconia [20] FGMs for biomedical applications.

Among the various AM methods for FGMs, Directed Energy Deposition (DED) holds an important place in the fabrication of metallic FGMs and finds extensive application in various industries for fabrication of high-degree precision components with advanced properties [11]. Furthermore, Laser-assisted Directed Energy Deposition (LDED) has been increasingly utilized as an advanced technique for fabricating FGMs where the processing parameters have substantial influence over microstructure and mechanical properties [16, 21].

To address the stated challenges in AM, Machine Learning (ML) techniques have emerged as a promising means for optimizing processing parameters, improving product quality, and detecting manufacturing defects [22]. ML techniques also offer a potential solution to challenges related to material design and development, whereas by establishing the complex relationships between composition, microstructure, process, and performance, these techniques enable the discovery of new materials with tunable and improved properties [23]. Notably, traditional topology optimization methods have been augmented by deep learning-based approaches, particularly in the optimization of composite structures [24]. Also, ML techniques involving in-situ monitoring have contributed to real-time defect detection, microstructure classification, and property prediction [5, 25–27].

Although multiple related literature review articles have covered various aspects of the implementation of ML methods in AM [22–25], to the best of our knowledge this is the first survey that focuses on ML methods for fabrication of FGMs. Considering the increased interest in FGMs applications and their potential for enhanced functionality across tasks and industries, this review article can benefit the interested community.

The paper is organized as follows. In Section 2, we provide a brief overview of studies related to FGMs fabrication, followed by reviewing works on employing ML in AM under Section 3, and a brief overview of ML methods in Section 4. Section 5 is the main section of the article, and it presents published studies from the literature related to the application of ML techniques for FGMs fabrication. Next, we discuss future directions and challenges in the application of ML for FGMs fabrication, and the last section concludes the paper.

## 2. ADDITIVE MANUFACTURING FOR FGMs FABRICATION

Directed energy deposition (DED) is a category of AM processes in which a feedstock material in the form of powder or wire is delivered to a substrate on which an energy source such as a laser beam, electron beam, or plasma/electric arc is simultaneously focused. The melted powder forms a melt pool and results in continuously depositing material, layer by layer. DED has several unique advantages compared to other AM processes, such as site-specific deposition, alloy design, and three-dimensional printing of com-plex shapes [28, 29]. Figure 1 presents a schematic of the DED process of FGMs [30].

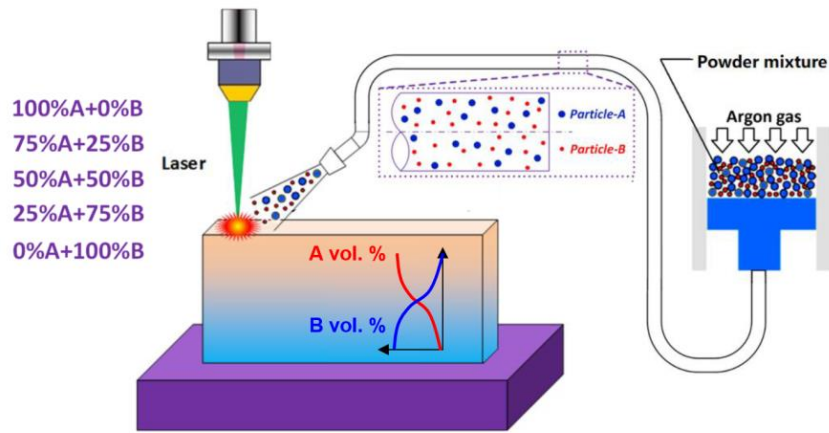

**Figure 1**. Visualization of the Directed Energy Deposition (DED) process of FGMs, showcasing the deposition of two materials onto the substrate via a laser-based energy source [30].

Laser-assisted DED (LDED) is an advanced AM technology for fabricating FGMs, characterized by differences in properties as the dimension varies. The overall properties of FGMs are unique and different from any of the individual materials that form them. A schematic illustration of the three key aspects of the formation of FGMs from Mahmoud et al. [31] is shown in Figure 2.

A DED system equipped with multiple powder feeders and nozzles can create FGMs by varying the powder flow rate from each feeder [32]. In the DED process, the micro-structure and mechanical properties of the parts can be largely affected by the processing parameters. The processing parameters also affect the deposition efficiency and mechanical properties of deposits [33]. Zhang et al. [21] introduced a DED fabrication process where a Cobalt-based alloy W50 was deposited on an H13 steel substrate. Since DED parameters have significant effects on parts characteristics, the processing parameters were optimized and used to fabricate samples for mechanical tests. Experimental results showed that tensile test samples had low residual stress, confirming that by optimizing DED processing parameters, a significant reduction of residual stresses can be realized in W50 coatings. Consequently, the authors were able to achieve crack-free W50 coatings by optimizing DED parameters without substrate pre-heating.

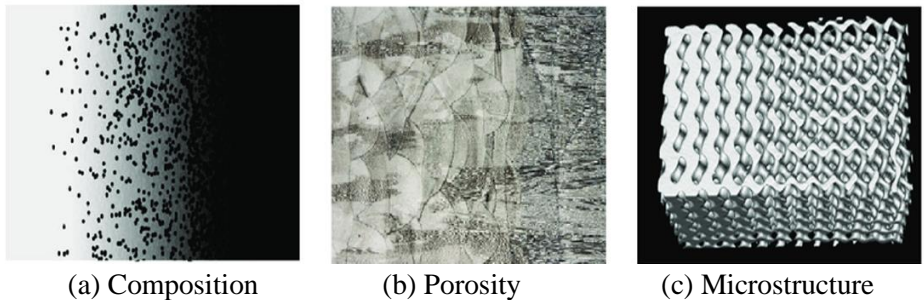

(a) Composition     (b) Porosity     (c) Microstructure

**Figure 2**. Illustration of three key aspects of the formation of Functionally Graded Materials (FGMs): a) composition variation, b) porosity distribution, and c) microstructural characteristics [31].

Another rapidly expanding area of research in AM is the utilization of location-dependent composition control for creating FGMs [34]. For instance, direct joining of copper and stainless steel (SS) is prone to cracking due to the very limited solubility be-tween Cu and Fe [35]. Zhang et al. [36] presented an innovative approach for fabrication of defect-free and crack-free FGMs using AM by introducing interlayers. The authors utilized a DED technique which involved controlled deposition of material through a focused energy source along with interlayers between SS and Cu material to improve bonding and mitigate the mismatch between their thermal and mechanical properties in achieving desirable residual stress.

Li et al. [16] developed a physics-modeling approach to simulate the fabrication of FGMs using DED. The main objective was to predict the thermal distribution, material composition, and molten pool dynamics of FGMs. The authors validated the developed models for FGMs comprising Stainless Steel 316L (SS316L) and Inconel 718 (IN718) parts with thin-wall structure. The proposed methodology involved multi-physics and multi-material modeling, considering heat transfer, fluid flow, and material interactions. The models employ conservation equations for momentum, energy, and mass, and incorporate material properties and composition ratios of SS316L and IN718 through a mixed-material model with volume fractions. The model simulated the DED process with different processing parameters while accounting for composition changes. Experimental validation encompassed fabricating SS316L/IN718 FGMs with thin-wall structure. The comparison of simulated and experimental outcomes demonstrated the model's ability to estimate material composition ratios and geometry in FGMs.

In another work, Li et al. [37] investigated the microstructural evolution, grain growth, and precipitation behaviors in FGMs with varying compositions of IN718 and SS316. Optical microscopy (OM) and electron backscatter diffraction (EBSD) were employed for microstructural characterization. OM employs visible light and a system of lenses to provide an overview of the material microstructure, including the grain size, morphology, and distribution of phases. EBSD is a scanning electron microscopy (SEM) technique, which uses the diffraction pattern of backscattered electrons from a focused electron beam to determine the detailed maps of crystal orientation and phase information within a material. EBSD provides information at a much higher resolution in comparison to OM. In the study by Li et al. [37], the OM observations revealed a transition from columnar to equiaxial grain shapes as the weight percent of IN718 in the FGM parts increased. This behavior was consistent with the EBSD mapping, which showed that the component with 75 wt% IN718 has the finest grain size, attributed to the effect of heat capacity on solidification and grain growth. The authors were able to fabricate FGMs from SS316 to Inconel 718 using powder-based LDED exhibiting a gradient microstructure without defects and distortion.

## 3. MACHINE LEARNING METHODS IN ADDITIVE MANUFACTURING

Over the last two decades, the design, processing, and production of AM parts have become more complex. Consequently, optimizing the AM process parameters has become more time-consuming and costly. As a result, ML methods have been increasingly applied in AM to reduce human effort and time [22]. ML-based tasks in AM encompass research for part design (e.g., to accelerate topology optimization and provide greater complexity in constraints, or generate lattice structures based on mechanical properties), process parameter optimization and process monitoring, and development of tools for general pro-duction planning [22]. ML models have also been developed to identify the manufacturability of components and dimensional deviations in AM parts [22].

In a recent survey of ML methods for material design of steel materials, Pan et al. [23] divided ML algorithms in materials science into four general categories: classification, clustering, regression, and probability estimation. For material property prediction and microstructure identification, ML algorithms that employ classification, regression, and clustering proved more beneficial, whereas probability estimation algorithms are more common in materials development. Also, Bayesian algorithms, genetic algorithms, and decision tree algorithms were used in prior works for optimizing process parameters. The authors emphasized the potential that ML offers for overcoming the complex physical mechanisms in materials and for the development of advanced steels [23].

Another survey paper on the use of ML in AM was recently published by Ciccone et al. [24]. The researchers selected 48 papers from the comprehensive collection of works and provided a systematic literature review to assess the potential of AI applications in optimizing AM processes and techniques. The review covers research for augmenting AI models with additional data to improve accuracy, implementing real-time solutions for quality control, integrating additional process parameters for more precise predictions, and creating comprehensive optimization approaches that combine multiple AI methods. Other reviewed topics include standardizing data reporting to address data scarcity, using AI in the design process for complex structures, predicting the structural strength of printed components, assessing material performance through microstructure analysis, optimizing tool paths for efficiency, and monitoring and controlling AM processes in real time.

Fu et al. [25] conducted a review of ML algorithms for defect detection in Laser-Based Additive Manufacturing (LBAM). The ML algorithms were applied to different types of image data to enhance the efficiency of the printing process. The authors considered several types of defects, such as porosity, which is a common issue in LBAM that affects product density. Incomplete fusion holes occur due to insufficient energy input during the LBAM process, whereas cracks can result from high laser energy input leading to high-temperature gradients and potential crack initiation during or after solidification. Various types of data have been used for defect detection and classification, including Melt Pool Thermal (MPT) data, Acoustic Signals (AE), optical/layer-wise images, photo-detector data, powder bed images, and MWIR images. The authors reviewed 48 papers with different case scenarios including varying material properties, fabrication methods, and defects. The most common method for supervised learning is Convolutional Neural Networks, whereas popular unsupervised learning methods include k-Nearest Neighbors, Deep Belief Networks, and Self Organized Maps algorithms. Reinforcement learning-based approaches have also been explored for efficient fault detection models. The article also discusses the importance of selecting suitable ML algorithms based on printing technologies, materials, and defect types.

## 4. OVERVIEW OF MACHINE LEARNING METHODS

ML is a subfield of Artificial Intelligence that focuses on developing algorithms that learn from data and make predictions on unseen data. ML algorithms are in general categorized into supervised and unsupervised learning algorithms. Supervised learning entails the application of an algorithm for processing a dataset consisting of training samples and labels. Generally, the labels can be discrete such as indicating class membership (e.g., presence or absence of a detrimental phase in an FGM) or continuous values (e.g., the level of porosity of an FGM). Unsupervised learning algorithms are developed to extract knowledge or insights about a dataset of training samples without labels, and often focus on identifying patterns or similarities in the training samples.

The following section provides a brief overview of common ML algorithms and methods. For a detailed description, the reader can refer to the following works [38, 39].

### 4.1 Unsupervised Learning Algorithms

Unsupervised learning comprises ML algorithms that learn from unlabeled datasets and involves tasks such as dimensionality reduction, clustering, and density estimation. Dimensionality reduction refers to reducing the number of input features in a dataset, and it is often applied when dealing with training samples having a large number of features. Common ML methods for dimensionality reduction include Principal Component Analysis (PCA) [40], Linear Discriminant Analysis (LDA) [41], and t-distributed Stochastic Neighbor Embedding (t-SNE) [42]. In addition, ANNs-based techniques have also been employed for dimensionality reduction. Clustering is used to group the training samples of a dataset into clusters based on the similarity of their features. Popular clustering methods include K-Means clustering [43], Mean Shift clustering [44], Density-Based Spatial Clustering of Applications with Noise (DBSCAN) [45], and Hierarchical Clustering [46]. Density Estimation entails estimating the probability density function of the features for the training samples, in order to understand the distribution of the dataset. ML methods for density estimation include Gaussian Mixture Models (GMMs) [47], Kernel Density Estimation (KDE) [48], and Probabilistic Graphical Models (PGMs) [49].

Unsupervised learning algorithms are less commonly used in FGMs fabrication, in comparison to supervised learning algorithms. However, several related works employ dimensionality reduction for feature selection as a preprocessing step before applying a supervised learning algorithm.

**4.2 Supervised Learning Algorithms**

Supervised learning algorithms are often divided into two main categories, classification and regression, based on whether the data labels are discrete or continuous. Classification algorithms employ discrete labels, and the objective typically is to classify the training samples into two or multiple classes. Regression algorithms use continuous labels, and the objective is to predict the value of a target variable for a given training sample.

Classification algorithms can be further categorized into several subgroups, based on the working principles upon which the algorithms predict the class membership. Numerical classifier ML algorithms are designed to approximate a mapping function between the input samples and the labels by optimizing an objective function. ML methods in this subgroup include Logistic Regression (LR) [50], Linear Classifier (LC) [51], Perceptron algorithm [52], Support Vector Machines (SVM) [53], and Artificial Neural Networks (ANNs) [39]. Probabilistic ML algorithms employ parametric probability distributions to model the mapping between the training samples and the labels, and include Naïve Bayes (NB) [54], Gaussian Discriminant Analysis (GDA) [38], Hidden Markov Models (HMMs) [55], and Probabilistic Graphical Models (PGMs) [38]. Instance-based non-parametric algorithms perform the learning task employing the individual data instances directly, without deriving model parameters. An example of an instance-based algorithm for classification is k-Nearest Neighbors (kNN) [56], and in addition, several regression algorithms have been developed based on the instance-based approach. Symbolic ML algorithms employ high-level symbolic representation of the training samples and perform the learning task based on logic and search via manipulation of symbols. Representative symbolic methods include Decision Trees (DT) [57] and Classification And Regression Trees (CART) [58]. Lastly, ensemble learning algorithms perform classification by aggregating the predictions by a collection of base learning algorithms. Commonly used ensemble methods include Random Forest (RF) [59], Gradient Boosting (GB) [60], Extreme Gradient Boosting (XGBoost) [61], and Bagging Ensemble (BA) [62].

Regression algorithms in ML can also be categorized into similar subgroups as classification algorithms, and even some methods can be used both for classification and regression learning tasks. Numerical regression methods based on optimizing an objective function include Linear Regression (LiR) [63], Polynomial Regression (PR) [63], Support Vector Regression (SVR) [64], and Artificial Neural Networks (ANNs). Instance-based non-parametric algorithms include Kernel Regression (KL) [65] and Local Regression (LoR) [66]. Symbolic ML algorithms include Decision Tree Regression (DTR), as well as most ensemble methods, such as Random Forest (RF) [59], Gradient Boosting (GB) [60], and Extreme Gradient Boosting (XGBoost) [61] can be employed for regression tasks.

### 4.3 Artificial Neural Networks

Artificial Neural Networks (ANNs) belong to the group of numerical optimization approaches, where the learning task is accomplished by minimizing an objective function between the training samples and target variables. ANNs use layers of computational units called neurons for learning the mapping function. The neurons process the values received from the preceding layer, apply an activation function, and pass the outputs to the successive layer in the network. The parameters of the network are learned during the training phase, which includes forward propagating the input features through the network followed by backward propagating the predicted values in order to minimize the objective function. Common objective functions with ANNs are Cross-Entropy for classification tasks and Mean-Squared Error for regression tasks.

ANNs with more than three layers are referred to as Deep NNs (DNNs), where the depth of the network refers to the number of hidden layers. Recent ANNs typically comprise several tens or hundreds of hidden layers containing millions or billions of learned parameters. A variety of different architectures of ANNs have been developed and applied to different tasks. Common architectures include Fully-Connected NNs (FCNNs) commonly referred to as Multi-Layer Perceptrons (MLPs) [39], Convolutional NNs (CNNs) [67], Recurrent NNs (RNNs) [68], Transformer NNs (TNNs) [69], Probabilistic Diffusion NNs (PDNN) [70], Graph NNs (GNNs) [71], and Generative Adversarial Networks (GANs) [72].

### 4.4 Reinforcement Learning

Reinforcement Learning (RL) [73] is another branch of ML, which is based on maximizing a reward function calculated via interaction with an environment. Unlike super-vised and unsupervised learning, RL relies on trial and error to discover the optimal actions that maximize the rewards. The learning process is guided by a policy, which defines the strategy for selecting actions based on the current state. An important element of RL is achieving a balance between exploration and exploitation, where exploration refers to trying new actions that can potentially lead to higher rewards, and exploitation in-volves selecting past actions that maximize the rewards.

One of the most used RL algorithms is Q-learning [74]. It is a value-based approach where a value function is iteratively learned to estimate the cumulative reward of taking an action while being in a specific state. Deep Q-Network (DQN) [75] is a variant of Q-learning that employs Deep NNs for approximating the value function, being particularly suited for complex environments with high-dimensional inputs. Other popular RL algorithms are Policy Gradient (PG) methods [76] that directly optimize the policy by calculating gradients with respect to the policy parameters. Proximal Policy Optimization (PPO) [77] is an example of a PG method that constrains the updates of the policy to prevent sudden changes to the policy. Actor-Critic RL algorithms [78] combine value-based and Policy Gradient methods by employing a value-function, i.e., critic, to evaluate the actions of the policy, i.e., actor.

A graphical overview of common ML methods [79] is provided in Figure 3.

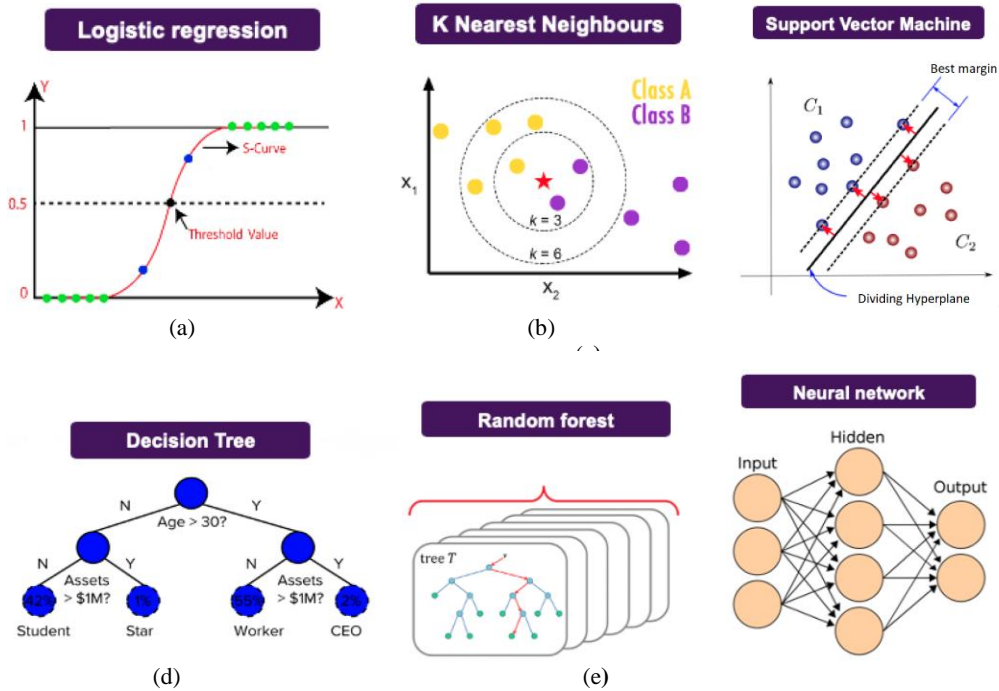

**Figure 3**. Common ML methods [79]. (a) Logistic regression – calculates parameters that map inputs through a logistic sigmoid function (the red curve in the figure), and classifies a sample based on a threshold value; (b) k-Nearest Neighbors – the class for a new sample (the star in the figure) is predicted based on the class labels of k closest samples; (c) Support Vector Machines – identifies a dividing hyperplane that separates the blue and red samples in the figure by finding the maximum margin between the hyperplane and the samples; (d) Decision Tree – classifies samples by asking a series of questions and splitting the data at each node based on selected threshold values; (e) Random Forest – aggregates the predictions by many Decision Trees to obtain a more robust model; (f) Artificial Neural Network – uses hidden layers to learn a mapping function between the training samples (inputs) and the target variables (outputs).

## 5. MACHINE LEARNING METHODS FOR FGMs FABRICATION

### 5.1 Parameters Optimization and Prediction of Material Properties

Kim et al. [80] explored the design optimization of FGM composite structures in comparison to traditional fiber-reinforced composites and introduced a new framework using the representative volume element (RVE) method and ML for the simultaneous de-sign of topology and fiber layout in FGMs. An Artificial Neural Network (ANN) was used to predict the effective material properties based on given design variables. The model was trained using data generated from RVE simulations, enabling quick and accurate predictions of material properties for different configurations. This approach accelerated the optimization process, making it computationally efficient. Additionally, constraints were imposed on the composite volume fraction and the fiber material volume fraction to ensure the practicality of the design. The proposed methodology was validated using examples with multi-load and 3D cases, and the results demonstrated that the optimized FGM design outperformed the traditional fiber-reinforced composite design in terms of structural compliance and stiffness.

Srinivasan et al. [81] developed an approach for coupling physics-based process modeling with ML to accelerate searching the AM processing space for suitable sets of processing parameters in Laser Powder Bed Fusion (LPBF). The methodology involved simulating thermal histories, reducing the dimensionality of thermal histories, and optimizing laser scan parameters. The approach was validated with Ti-6Al-4V alloy for fabrication of parts with simple and complex geometries. The objective was to enhance the mechanical properties and performance of manufactured parts by minimizing the thermal heterogeneity. A calibrated thermal model was developed that accurately simulates the temperature evolution during the LPBF process. The thermal model served as a tool for predicting the thermal histories of fabricated components and understanding how processing parameters influence the material properties. The model considers factors such as laser power, laser velocity, and hatch spacing in the LPBF process. Principal Component Analysis (PCA) was applied to reduce the dimensionality of the vectors to a lower-dimensional space. Afterward, a Density-Based Spatial Clustering of Applications with Noise (DBSCAN) was utilized for modeling the variations in thermal histories. By optimizing scan parameters locally, the researchers achieved more uniform part processing and significantly reduced the variation in thermal histories, which led to improved mechanical properties and enhanced overall performance of the printed components, as well as reduced heterogeneity in LPBF.

Dharmadhikari et al. [8] proposed a model-free Q-learning Reinforcement Learning algorithm for process parameter optimization of melt pool depth within the power-velocity (P-V) domain of a candidate LDED process. The experiments were performed by varying the power and velocity of the LDED system and they were afterward used to generate a process map of the melt pool depth. The results from the different P-V combinations in the generated process map were used as a validation tool to verify the optimal solutions returned by the algorithm. The experimental results proved the effectiveness of the model in predicting optimal combinations of power and velocity.

To establish the relationship between the build height and density of DED samples and fabricate defect-free and high-degree precision components, Narayana et al. [11] de-signed an ANN to optimize the processing parameters for DED. The ANN model was beneficial in reducing the number of required experimental trials, and thereby minimizing material waste and fabrication time. To optimize the model parameters of DED, the laser power, scan speed, powder feed rate, and layer thickness were considered as inputs, and the density and build height of the samples were the outputs of the model. To examine the processing parameter influence, the authors used 3D plots (process maps) based on the ANN predictions displaying the relationship between the feed rate and power, and feed rate with thickness for density and build height separately, highlighting the optimal regions in each process map. To validate the ANN results, four new sets of experiments were conducted by selecting various process parameters from the designed process map. The results demonstrated that the model predictions for combinations of height and density matched well with achieving the optimal region in the process maps.

Alcunte et al. [82] investigated the fatigue properties of fabricated Polylactic Acid (PLA) and Thermoplastic Polyurethane (TPU) specimens under uniaxial loading. The specimens were fabricated using Multi-Material Additive Manufacturing (MMAM), which allows for customized variations in material characteristics, such as mechanical strength, thermal conductivity, and stiffness. A ZMorph single-nozzle dual-material 3D printer was used to manufacture the multi-material structures. The study utilized multiple ML models, including Random Forest (RF), Support Vector Machine (SVM) classifier, and Artificial Neural Network (ANN). The data for model training was obtained from experiments conducted using a Testresources 810E4 load frame with a 15 KN load capacity. The experiments showed that the RF model outperformed SVC and ANN in predicting fatigue cycles of polymeric FGMs, achieving the lowest RMSE values on the test dataset.

Raturi et al. [83] studied the mechanical characteristics of porous FGMs, with a particular focus on an even distribution of porosity for FGM plates. An important area of re-search in this context is free vibration analysis which examines the natural frequencies of a material without external support. The FGM plate in the study was designed using stainless steel SS304 alloy as the first material and silicon nitride (Si3N4) as the second material. To evaluate the fundamental natural frequencies, a Finite Element Method (FEM) was used with input parameters comprising elastic and shear moduli along the direction of fibers, Poisson's ratio, mass density, and thickness. The authors employed multiple ML models including Linear Regression (LR), Gaussian Process Regression (GPR), Artificial Neural Network (ANN), and Support Vector Machine (SVM). The experimental validation revealed that LR and GPR provided more accurate predictions when compared to SVM and ANN.

Sulaiman et al. [84] investigated the efficiency of FGMs annular fins in heat transfer applications, specifically for analyzing heat transfer and temperature distribution in heat exchangers, electronic cooling systems, and power generation equipment. The main purpose of adding an annular fin to an object is to increase the surface area in contact with the surrounding fluid, enhancing convective heat transfer. With a growing demand for fins, it is important to consider thermal parameters such as thermal conductivity, emissivity, and heat transfer coefficient. An ANN model was employed, where a nonlinear ordinary differential equation was used to simulate ring fin heat transfer and generate data. The authors used 2,001 individual data points and a loss function based on the Broyden–Fletcher–Goldfarb–Shanno (BFGS) algorithm to train the model. The analysis showed that the thermo-geometric parameter significantly impacts local temperatures in annular fins, with lower values resulting in faster conductive heat transfer and higher local temperatures due to the inverse correlation with thermal conductivity.

Wasmer et al. [85] conducted a study to develop a system for real-time monitoring of FGMs fabrication by an LDED process. The selected materials included Titanium (due to its stable solid α and β phases and biocompatibility for medical applications) and Niobium. The study produced parts ranging from pure Titanium to pure Niobium with intermediate mixtures, listed as 100%Ti, 58%Ti42%Nb, 37%Ti63%Nb, and 100%Nb. Re-al-time monitoring during the fabrication of FGMs samples is crucial for ensuring part quality and consistency, defect detection, and improving the efficiency of the AM process. Commonly used sensors for real-time monitoring include vision cameras, thermal cameras, optical sensors, and acoustic emission sensors. In this study, for real-time monitoring of the AM process, the authors utilized three types of sensors that were installed on an industrial LDED machine: a microphone, optical emission spectroscopy (OES) sensor, and a vision camera. The microphone captured the acoustic emission (AE) signals emitted during the fabrication process, the OES sensor gathered optical spectra information, and the vision camera captured images for monitoring the process zone dynamics. In addition, the fabricated parts were analyzed post-hoc using optical microscopy (OM), scanning electron microscopy (SEM), and energy-dispersive X-ray spectroscopy (EDX) to establish the ground truth for evaluating the ML models. Whereas OM provides low-resolution morphological information and phase identification, SEM provides high-resolution imaging with detailed microstructure information down to nanometers. EDX is typically combined with SEM and employs the emitted X-ray energies from the high-energy electrons for determining the elemental composition, creating maps of elemental distribution across FGM samples, and differentiating phases based on their elemental composition. Hence, EDX augmented the information obtained by SEM with elemental and compositional analysis. The authors implemented several ML models to classify the data collected online during the LDED process, including Linear Discriminant Analysis (LDA), Logistic Regression (LR), Support Vector Machine with Linear Kernel (LSVM), Support Vector Machine with RBF Kernel (SVM), k-Nearest Neighbors (kNN), Multi-Layer Perceptron (MLP), and Random Forest (RF). The MLP model achieved the highest average accuracy of 96.38%. Further, training the models with data collected with the OES sensor achieved high classification accuracies across all ML models, indicating its potential for online monitoring of FGMs with varying process regimes. The signals from the OES sensor effectively distinguished between different chemical compositions, even when tilted at an extreme angle. The collected sensory data was helpful for differentiating between various process regimes such as conduction mode and different types of lack of fusion pores, demonstrating its capability to detect details in the AM process.

Table 1 summarizes the investigated works for process parameter optimization and prediction of material properties in AM using ML.

**Table 1.** Investigated literature regarding the application of ML in process parameter optimization and prediction of material properties in additive manufacturing.

| Title | Drawback | ML method | Data available | Investigated materials | Predicted parameters/properties |
|---|---|---|---|---|---|
| Kim et al. [80] | N/A | ANNs | N/A | N/A | Elasticity, stiffness |
| Srinivasan et al. [81] | Not suitable for complicated geometries | PCA, DBSCAN | Yes | Ti-6Al-4 V | Laser power, laser velocity, hatch spacing |
| Dharmadhikari et al. [8] | Focus on optimizing geometric tolerance, limited space action | Q-learning | Yes, upon request | N/A | Power and velocity |
| Narayana et al. [11] | N/A | ANN | Yes | Ti-6Al-4V | Scan speed, powder feed rate, and layer thickness |
| Alcunte et al. [82] | Need to improve accuracy across different stress levels | RF, SVC, ANN | Yes, upon request | PLA, TPU | Fatigue life |
| Raturi et al. [83] | N/A | LR, GPR, ANN, SVM | Yes, upon request | SUS304, Si3N4 | Natural frequencies of porous FGMs |
| Sulaiman et al. [84] | Numerical methods implementation | ANN-BFGS | Yes, upon request | N/A | Heat transfer and temperature distribution of ring fins |
| Wasmer et al. [85] | OES Challenges, Feature Reduction Impact | LDA, LR, SVM, kNN, MLP, RF | Yes | Titanium, Niobium | Lack-of-fusion porosity, real-time monitoring |

## 5.2 Classification of Phases in Microstructure

Challenges that can cause the failure of FGM parts/components include mismatched lattices, differences in thermal expansion coefficients, and difficulty in optimizing printing parameters for multiple materials simultaneously. Eliseeva [86] introduced a novel computational design method for FGMs that plans the compositional gradients such that detrimental phases are avoided. This method enables FGMs gradient paths to be planned in high-dimensional spaces and optimized with respect to an objective function. The method consists of two major steps. First, an ML technique is used to map regions in phase diagrams comprising the multi-dimensional composition-temperature space that may contain detrimental phases over a range of temperatures. This is followed by the use of a path-planning algorithm adapted from robotics to devise a compositional gradient path that optimizes a selected objective function. Regions of detrimental phases identified during the first step are considered to be obstacles to be avoided during the path planning step. This approach resulted in gradient pathways that are more robust to manufacturing temperatures, cooling rates, or post-fabrication heat treatment. Gradient samples were successfully fabricated using a multi-material DED process with optimized processing parameters.

Galvan et al. [87] worked on determining thermodynamic conditions for a desired phase state where the goal was to find a generalized inverse phase stability solution for material discovery. The paper investigated both Gradient-Improvement Particle Swarm Optimization (GIPSP) and the Ensemble-Disjoint Set Discrimination (EDSD) algorithms, which are designed to address the challenges of complex and multidimensional Constraint Satisfaction Problems (CSPs) encountered in the domain of material design to tackle the highly nonlinear and discontinuous search space of the generalized inverse phase stability problem. The authors focused on optimizing material properties by identifying thermodynamic conditions that satisfy specific phase state constraints in the Fe-Ti binary alloy system. The results demonstrated that the GIPSP algorithm consistently outperformed the EDSD algorithm in terms of precision and recall. The GIPSP algorithm exhibited better accuracy in approximating the true constraint boundary, especially in cases where the satisfactory region was small relative to the search space. The EDSD algorithm, however, had significant difficulty in approximating the solution for some cases which might be attributed to the limitations of the support vector data description technique used in EDSD.

Linearly graded composition in FGMs often leads to undesirable phases and cracking. To address this problem, Kirk et al. [88] focused on designing and optimizing FGMs with monotonic property profiles. To this end, the authors introduced a methodology that enables efficient identification of optimal paths in the composition space of FGMs, while ensuring that these paths exhibit monotonic variations in thermal expansion. A novel cost function was proposed that prioritizes paths with monotonic property variations and it integrates several metrics, including the lack of increase (LOI) and lack of decrease (LOD), which were derived from the calculation of the positive and negative parts of the first derivative of the property function over the composition space. The researchers used a Rap-idly-exploring Random Tree with Final Node (RRT*FN) algorithm, a sampling-based planner known for its efficiency in exploring high-dimensional spaces and constructing paths between the initial and target compositions. The objective was to ensure that the proposed algorithm finds paths that are both optimal in length and possess the desired monotonic variations in thermal expansion. By carefully controlling the deposition rate and adjusting constraints, the study demonstrated that paths with smooth property pro-files can be achieved, even in regions with steep changes in property values.

Table 2 summarizes the different investigated studies for avoiding detrimental phases in AM using ML.

**Table 2.** Investigated literature regarding the application of ML in avoiding detrimental phases in additive manufacturing.

| Title | Drawback | ML method | Data available | Investigated materials |
| --- | --- | --- | --- | --- |
| Eliseeva et al. [86] | N/A | Path planning | Cannot be shared | SS316L, chromium, nickel (Fe-Ni-Cr) |
| Galvan et al. [87] | Limited dimensionality | GIPSP, EDSD | Yes - ThermoCalc | Fe-Ti binary alloy compositions |
| Kirk et al. [88] | Inability to optimize multiple properties simultaneously | RRT*FN | Yes, upon request | Fe-Co-Cr system |

## 5.3 Detection of Defects

The quality of manufactured parts is a major issue in AM since defects negatively influence the structural integrity of the parts and because of the complex physics involved in the AM process. To address challenges in qualifying and certifying parts due to the variability of the manufacturing process, Mojumder et al. [89] introduced an ML method for predicting LOF (lack-of-fusion) porosity and understanding its relationship to the processing conditions. Identifying the relationship between process parameters and LOF porosity is a challenging problem due to the high dimensionality of the process design space and the large number of simulations or experiments required to evaluate all processing conditions [2, 3]. To establish the relationship between processing conditions and LOF porosity, the authors proposed an ANN architecture that employed symbolic regression and predicted the LOF porosity using a physics-based thermo-fluid model. Specifically, the porosity was predicted using the thermo-fluid model and afterward, the predicted porosity data for different processing conditions were used to show the efficacy of an active learning framework to reduce the number of simulations for effectively mapping the process design space. Finally, the samples from active learning were used to predict the processing parameters and porosity relationship. The limitations of the proposed model are that it does not consider multiphase or multi-species flow and assumes that materials are fully melted when their temperature exceeds the melting point, leading to a simplified mechanism of porosity formation. Furthermore, the material properties used in the research were taken from the published literature and may vary considerably with temperature, which was assumed to be constant.

Melt pool monitoring is often used to ensure the quality of AM parts. For instance, Akbari et al. [9] developed an ANN-based method called MeltpoolNet that enables predicting the geometry and defect type of the melt pool in DED. The data points were collected from experiments using various alloys and processing parameters, and the model consists of a regression and a classification branch. The regression branch predicts the depth, width, and length of the melt pool, and the classification branch predicts the defect mode of the melt pool. The researchers formulated a constrained optimization problem using dimensional analysis and nonlinear regression to identify equations that relate the process parameters to the melt pool depth, width, and length. Good agreement was identified with the approximations derived from the theoretical Rosenthal equation. In this study, ANN, Gradient Boosting, and Random Forest outperformed other ML models applied to this task.

Van Houtum et al. [90] proposed the Adaptive Weighted Uncertainty Sampling (AWUS) method for predicting the quality of the melt pool in DED. AWUS is an active learning strategy designed to balance random exploration and uncertainty-based exploration using model change. The study also used in-situ acquired imaging data to differentiate between different quality levels, based on the extraction of features from the thermal images. The feature extraction combined gradient and intensity information, capturing the melt-pool boundary's visibility. The extracted features were used to train various classification models, including Logistic Regression (LR), Support Vector Machines (SVM), Random Forest (RF), and Naive Bayes (NB). The results demonstrated the effectiveness of the proposed AWUS active learning strategy, which outperformed state-of-the-art methods by significantly reducing the annotations needed for training across diverse datasets, classifiers, and batch sizes. Furthermore, the process quality classification method for DED exhibited promising performance, achieving a median F1-macro score of over 90% across distinct datasets acquired from various DED machines, emphasizing its practical suitability for real-world applications.

The presence of porosity in AM parts can significantly influence their mechanical properties and overall performance, making it a crucial property for AM parts. Eliseeva et al. [91] introduced an ML method to identify porosity in parts produced through LPBF AM processes. The proposed methodology involves developing a transfer function that connects location-specific processing information with microstructural features related to porosity. The approach utilizes an ML classifier to link attributes of the time-temperature history to the presence or absence of porosity in specific locations of the fabricated part. A 3D model of the printed part and its microstructure were analyzed to identify important features. Simultaneously, a digital twin of the AM process was generated to capture location-specific process representations. The methodology achieved good results in predicting porosity locations in AM parts, and it can be further extended to identify other thermally-linked microstructural features. The main limitation of this approach is the need for additional data to improve the effectiveness of the classifier, especially given the relatively rare occurrence of porosity events.

In Hespeler et al. [5] the authors emphasized the role of quality control in AM in ensuring desired mechanical properties and avoiding defects. The authors used two metal AM samples, one representing high-quality layers and one with low-quality layers. During the printing of the samples, in-situ process data were collected, including laser power, melt pool size, scan line energy density, powder concentration, gas, and temperature. The data was then preprocessed and standardized, and feature analysis techniques were applied, including Random Forest (RF), Decision Tree (DT), and XGBoost. The results revealed that laser power, melt pool size, and scan line energy density were the three most significant parameters affecting layer quality in metal AM. Afterward, a Convolutional Neural Network (CNN) model was trained on the collected in-situ data and was applied for layer-wise classification of the fabricated parts as acceptable or unacceptable. The layer-wise evaluation of the AM process unveiled that the unacceptable layers were more concentrated in the early stages of the build process, while the acceptable layers displayed a more uniform distribution. Although the classifier achieved high classification accuracy it did face overfitting, and the authors proposed expanding the dataset and exploring regularization techniques to enhance the model's robustness.

The work by Narayana et al. [11] involves identification and characterization of lack-of-fusion defects along the layer boundaries in the printed samples. The authors examined macroscopic images and microstructural features, including columnar grain size and phase constituents, to gain insights into the defect formation mechanisms under different process parameters. The detection of such defects is crucial for understanding the impact of laser power, scan speed, layer thickness, and powder feed rate on the quality of the printed alloy. By employing an Artificial Neural Network (ANN) model, the paper not only contributed to the optimization of process parameters but also investigated defect-related observations, aiming to mitigate or eliminate defects and enhance the overall quality of the Ti-6Al-4V alloy produced through DED.

Table 3 summarizes the investigated works for detection of defects in AM using ML.

**Table 3.** Investigated literature regarding the application of ML for detection of defects in additive manufacturing.

| Title | Drawback | ML method | Data available | Investigated materials |
|---|---|---|---|---|
| Narayana et al. [11] | N/A | ANN | Yes | Ti-6Al-4V |
| Mojumder et al. [89] | Uncertainty in porosity prediction - limited to LOF porosity detection | Active learning, NN | Yes | Ti-6Al-4V |
| Akbari et al. [9] | N/A | ANN, GB, RF | Yes | N/A |
| Eliseeva et al. [91] | N/A | SVM, KNN, Decision Tree | N/A | Ti-6Al-4V |
| Van Houtum et al. [90] | N/A | LR, SVM, RF, NB | Yes | N/A |
| Hespeler et al. [5] | Overfitting | RF, DT, XGBoost, CNN | Not public | N/A |

### 5.4 Discovery of New Materials

Discovery of new materials is a more challenging problem in comparison to predicting material properties, classification of phases, or defect detection. We can expect more work on this topic in the coming years as more data becomes available and the research efforts build up.

Functionally graded lattice structures (FGLSs) are intricate 3D forms with diverse material properties, potentially offering vast engineering and materials science applications. Veloso et al. [92] categorized lattice generation methods into Generative Design, Topology Optimization, Machine Learning, Genetic Algorithms, and Simulation-driven approaches. The paper discusses the utilization of ML for producing lattice structures with specific mechanical properties and compression ratios, alongside exploring genetic algorithms for optimization. Various research studies reviewed demonstrate the efficacy of these methods in generating tailored lattice structures.

Li et al. [93] employed inverse design to discover optimal structures for phononic crystals with the goal of controlling the light flow at the nano-scale. Phononic crystals are composite materials characterized by their periodicity which leads to unique properties and band gaps where waves cannot propagate. The unit cell of a nano-scale phononic FGM porous beam consists of two FGM porous cores labeled Core I and Core II. The porosity is symmetrically distributed relative to the mid-plane of the phononic crystal (PnC). To account for size effects at the nano-scale, Nonlocal Strain Gradient Theory (NSGT) is used considering both nonlocal stress and strain gradient stress. This theory aids in predicting complex phenomena like stiffness-softening or hardening due to size effects. A simplified form of the constitutive equation which includes the total stress tensor, classical strain, and strain gradient tensor, is provided. This equation reflects the length and thickness-related size effects. The displacement field is described using the Euler Beam assumption which simplifies the analysis by considering the beam's deflection in terms of its curvature. The mechanical properties of FGMs with porosity are determined using a modified power-law model, which accounts for continuously variable material properties and the effect of porosity. The authors derived the governing equations for the dynamic motion of phononic FGM porous nanobeams using Hamilton's principle. This principle requires the variations of the total potential energy, kinetic energy, and virtual work done by external forces to be zero. Designing phononic nanobeams that can exhibit specific band gaps is challenging due to the nonuniqueness of the problem as different structures can result in the same band gaps. To address this issue, the authors introduce the Probabilistic Tandem Network (PTN) model as a solution. This deep learning framework is de-signed to tackle the one-to-many problem in the inverse design of PnCs. It consists of an Inverse Neural Network (INN), Gaussian sampling, and a Forward Neural Network (FNN). The INN encodes band gaps into a latent space and the FNN decodes candidate structures from the latent space back to band gaps. The INN maps band gaps to a Gaussian distribution characterized by mean and standard deviation, which allows for the generation of multiple candidate structures. The deep learning models used for the inverse design of PnCs include Conditional Generative Adversarial Neural Network (CGAN), Conditional Variational Autoencoder (CVAE), Denoising Diffusion Probabilistic Model (DDPM), Denoising Diffusion Implicit Models (DDIM), Deep Neural Network (DNN), and Tandem Neural Network (TNN). A comparison of the accuracy and diversity performance between all the models is conducted to distinguish the inverse design model that performs best, with PTN outperforming the others. The authors concluded that both the analytical approach developed by the NSGT and the PTN model, based on the probabilistic super-vised learning manner can significantly contribute to the field of band gap engineering at a smaller scale.

Table 4 summarizes the investigated works for the discovery of new materials in AM using ML.

**Table 4.** Investigated literature regarding the application of ML for discovery of new materials in additive manufacturing.

| Title | Drawback | ML method | Data available | Investigated materials |
|---|---|---|---|---|
| Veloso et al. [92] | ML requires large amounts of data and expertise | Reviews ML methods, Genetic algorithm | Yes | N/A |
| Li et al. [93] | Design of PnCs at nanoscale | CVAE, DDPM, DDIM, DNN, TNN, PTN | Yes, upon request | Phononic Crystals |

## 5.5 Summary of Used ML Methods for FGMs Fabrication

Table 5 lists the ML methods described in this article on the literature review of FGMs fabrication, and provides the used method abbreviations in the text, the referenced works, and their advantages and disadvantages. Implementation of most conventional ML methods is available in various open-source and proprietary libraries, among which the most popular library is scikit-learn. Also, many related libraries provide implementations of ANNs, with the most popular libraries being Keras-TensorFlow and PyTorch.

**Table 5.** List of Machine Learning methods used in the reviewed works in this study.

| ML Method | Abbreviation | Referenced works | Advantages | Disadvantages |
|---|---|---|---|---|
| Artificial Neural Networks | ANNs | Kim et al. [80], Narayana et al. [11], Alcunte et al. [82], Raturi et al. [83], Sulaiman et al. [84], Wasmer et al. [85], Mojumder et al. [89], Akbari et al. [9], Li et al. [93] | Capture complex non-linear relationships, automatic feature extraction, can be used with various learning tasks | Require large datasets for training, computationally expensive, difficult to interpret |
| Logistic Regression | LR | Wasmer et al. [85], Van Houtum et al. [90] | Simple implementation, fast training | Assume linear relationship between data features and targets |
| Support Vector Machines | SVM | Alcunte et al. [82], Raturi et al. [83], Wasmer et al. [85], Galvan et al. [87], Van Houtum et al. [90] | Effective for high-dimensional data samples, can use different kernel functions | Computationally expensive, require finetuning of kernel type and parameters |
| Naïve Bayes | NB | Van Houtum et al. [90] | Simple implementation, easy to understand | Assume independence among the data features |
| $k$-Nearest Neighbors | $k$NN | Wasmer et al. [85] | Simple implementation, easy to understand | Requires storing the entire dataset for inference |
| Decision Trees | DT | Kirk et al. [88], Hespeler et al. [5] | Capture complex relationships, interpretable decisions | Prone to overfitting, training instability |
| Random Forest | RF | Alcunte et al. [82], Wasmer et al. [85], Akbari et al. [9], Van Houtum et al. [90] | Capture complex relationship, reduces overfitting of DT | Computationally expensive, less interpretable than DT |
| Gradient Boosting | GB | Akbari et al. [9] | High predictive accuracy for tabular data, interpretable decisions | Computationally expensive |
| Extreme Gradient Boosting | XGBoost | Hespeler et al. [5] | High predictive accuracy for tabular data, efficiency, interpretable decisions | Computationally expensive, requires parameter finetuning |
| Principal Component Analysis | PCA | Srinivasan et al. [81] | Suitable for dimensionality reduction and feature selection | Information loss during dimensionality reduction, assume linear relationship between features |
| Linear Discriminant Analysis | LDA | Wasmer et al. [85] | Suitable for dimensionality reduction and feature selection | Sensitive to outliers, assumes normal distribution of features |
| Density-Based Spatial Clustering of Applications with Noise | DBSCAN | Srinivasan et al. [81] | Can find clusters of arbitrary shape, can handle noise and outliers | Less effective for high-dimensional data, sensitive to parameter selection |
| Genetic Algorithm | GA | Veloso et al. [92] | Applicable for optimization of different problems, including non-linear functions | Computationally expensive, no guarantee of finding optimal solution |
| Linear Regression | LiR | Raturi et al. [83] | Simple implementation, fast training | Assumes linear relationship between features, sensitive to outliers |
| Gaussian Process Regression | GPR | Raturi et al. [83] | Capture complex, non-linear relationships, provides uncertainty estimates | Slow training with large datasets, require careful parameter finetuning |
| Convolutional Neural Networks | CNNs | Hespeler et al. [5] | Suitable for image processing, robust to variations in translations | Require large number of data samples, computationally expensive |
| Probabilistic Diffusion Neural Network | PDNN | Li et al. [93] | Provide probabilistic predictions, provide uncertainty estimates | Complex to implement and train, computationally expensive |
| Generative Adversarial Networks | GAN | Li et al. [93] | Suitable for different generative tasks, can create realistic data samples | Difficult and unstable to train, can suffer from mode collapse |
| Q-learning | QL | Dharmadhikari et al. [8] | Model-free RL approach, simple to implement | Struggles with large state-action spaces, can be sample inefficient |

## 6. FUTURE DIRECTIONS IN ML FOR FGMs

In the domain of AM for FGMs, the integration of ML has shown considerable promise, as evidenced by various research directions highlighted in the existing literature. However, several challenges remain, and promising future research directions emerge to address these issues.

One critical future direction involves the effective utilization of available data, especially from capturing parameter-process maps and microstructure-property maps. Exploiting in situ images and acoustic emissions could further enhance the performance of ML models, providing a comprehensive understanding of the complex relationships be-ween processing parameters and material properties [95, 96]. Addressing the challenge of data labeling in supervised learning could be mitigated by active learning strategies to facilitate more efficient acquisition and utilization of labeled data [32]. By allowing models to interactively query and label new data points during training, active learning minimizes the time, cost, and human labor associated with traditional labeling methods, thus enhancing the overall effectiveness of ML applications in AM. Similarly, the implementation of Reinforcement Learning strategies for continuous optimization of the processing parameters during FGMs fabrication [8] has the potential to provide improved control over the manufacturing process.

Uncertainty quantification in ML applications for AM presents another important research direction. Incorporating epistemic uncertainty into ML models, especially in regression tasks using methods like Gaussian Processes and Bayesian Neural Networks can provide important information about the confidence intervals of the predictions besides single point estimates [96]. Further exploration of uncertainty quantification procedures in the context of ML applications for AM of FGMs can contribute to improved robustness of the manufacturing process.

Transfer learning, identified as a valuable approach in the optimization of ML models for different AM processes and materials [32], offers an alternative to retraining models for each learning task. Future research should explore the potential of transfer learning techniques for facilitating the learning across different machines, materials, and part de-signs, thereby enabling adaptive and rapid model development.

Lightweight computing models for low latency are an important area for future development [42], due to the importance of addressing challenges in real-time monitoring and control scenarios. Innovations in creating efficient ML models that can ensure low latency and responsiveness are significant in the context of AM for FGMs.

In the context of AM for FGMs, cybersecurity is a pressing concern [22], as the evolving AM landscape introduces vulnerabilities, particularly in intellectual property (IP) protection. To address this, robust cybersecurity measures, including secure communication protocols and encryption, are essential. Developing industry-wide standards is also crucial for ensuring the integrity and safety of AM processes. While not directly related to ML, addressing cybersecurity is integral for the trustworthy and secure adoption of AM technologies.

## 7. CONCLUSION

In summary, this document surveyed recent papers on ML-based optimization in AM. The introduction highlighted the transformative impact of AM on production processes and emphasized its capacity for customization, complex design fabrication, and waste reduction. The methods section focused on ML for DED and for fabricating FGMs, in particular. Previous research enabled the prediction of thermal distributions, molten pool dynamics, and material composition, which are all crucial in shaping the properties of FGMs. ML can potentially offer solutions to the challenges of AM, as it provides algorithms for classification, regression, clustering, and probability estimation, each contributing to distinct aspects of AM. Furthermore, by providing insights into defect identification, parameter optimization, and real-time monitoring, ML can significantly enhance AM processes. The studies reviewed in this article include, but are not limited to, defect prediction, process quality classification, and the design of FGMs. The convergence of advanced fabrication techniques and data-driven methodologies is expected to gain importance in the coming years and offer novel manufacturing possibilities.